\documentclass{article}

\usepackage[utf8]{inputenc}
\usepackage[english]{babel}
\usepackage[letterpaper,top=2cm,bottom=2cm,left=3cm,right=3cm,marginparwidth=1.75cm]{geometry}
\usepackage{amsmath, amssymb}
\usepackage[colorlinks=true,allcolors=blue]{hyperref}
\usepackage{subcaption}  
\usepackage[utf8]{inputenc}
\usepackage{textgreek}
\usepackage{booktabs,multirow,siunitx}
\usepackage{microtype}

\usepackage{graphicx}          
\usepackage{tikz}
\usetikzlibrary{arrows.meta, positioning, matrix}
\usepackage{adjustbox}

\title{Transformer-Based Sleep Stage Classification Enhanced by Clinical Information}
\author{
    Woosuk Chung$^{1,2,*}$, Seokwoo Hong$^{1}$, Wonhyeok Lee$^{1}$, Sangyoon Bae$^{1,2}$\\ [0.4em]
  \small $^{1}$Risorius Inc., Seoul, Republic of Korea\\
  \small $^{2}$Department of Medicine, Seoul National University College of Medicine, Seoul, Republic of Korea\\
  \small $^{*}$Address for correspondence: tang58tang58@snu.ac.kr
}
\date{\today}

\begin{document}
\maketitle

\begin{abstract}
Manual sleep staging from polysomnography (PSG) is labor-intensive and prone to inter-scorer variability. While recent deep learning models have advanced automated staging, most rely solely on raw PSG signals and neglect contextual cues used by human experts. We propose a two-stage architecture that combines a Transformer-based per-epoch encoder with a 1D CNN aggregator, and systematically investigates the effect of incorporating explicit context: subject-level clinical metadata (age, sex, BMI) and per-epoch expert event annotations (apneas, desaturations, arousals, periodic breathing). Using the Sleep Heart Health Study (SHHS) cohort (n=8,357), we demonstrate that contextual fusion substantially improves staging accuracy. Compared to a PSG-only baseline (macro-F1 0.7745, micro-F1 0.8774), our final model achieves macro-F1 0.8031 and micro-F1 0.9051, with event annotations contributing the largest gains. Notably, feature fusion outperforms multi-task alternatives that predict the same auxiliary labels. These results highlight that augmenting learned representations with clinically meaningful features enhances both performance and interpretability, without modifying the PSG montage or requiring additional sensors. Our findings support a practical and scalable path toward context-aware, expert-aligned sleep staging systems.
\end{abstract}

\section{Introduction}

Polysomnography (PSG) is known as the gold standard in the clinical evaluation of sleep architecture and the diagnosis of various sleep disorders. According to the American Academy of Sleep Medicine (AASM), sleeping status is scored in 30-second epochs into Wake, N1, N2, N3 and REM stages based on PSG recordings including electroencephalogram (EEG), electrooculogram (EOG), electromyogram (EMG), and cardiorespiratory signals \cite{AASMManual2025,GriggDamberger2012}. Reliable identification of sleep stages is essential to diagnose and classify the severity of conditions such as obstructive sleep apnea (OSA), hypersomnia, parasomnia, and insomnia, and serves as the basis for many subsequent analytical and therapeutic processes \cite{StatPearlsSleepStudy}.

Nonetheless, manual sleep scoring remains both labor-intensive and subject to variability. Large-scale investigations have consistently shown inter-scorer concordance rates of approximately 80–85\%, with most discrepancies occurring in the N1 and N3 stages \cite{Rosenberg2013,younes2016staging}. In practical settings, the visual annotation of a single overnight PSG recording generally requires one to two hours of expert review \cite{Malhotra2013,Rayan2024}. Recent studies demonstrate that automated algorithms can shorten processing time by nearly two orders of magnitude while maintaining agreement levels comparable to human scorers across major PSG metrics \cite{Choo2023Benchmarking}. Therefore, the need for automated analytic approaches with high accuracy and interpretability is emerging.

Deep learning (DL) has greatly advanced automated sleep staging. Early end-to-end convolutional models directly learned temporal representations from raw or time–frequency PSG signals \cite{Chambon2018}. Later sequence-based frameworks, such as hierarchical RNNs and attention models in SeqSleepNet, captured longer temporal dependencies \cite{Phan2019SeqSleepNet}. Encoder–decoder architectures like U-Time enabled efficient full-night segmentation \cite{Perslev2019UTime} and scaled effectively to large heterogeneous cohorts, as shown in U-Sleep \cite{Vallat2021}. Attention mechanisms further improved intra- and inter-epoch modeling, exemplified by AttnSleep \cite{Eldele2021}.

Recently, transformer architectures have become state-of-the-art for PSG time-series, providing sequence-to-sequence staging, uncertainty estimation \cite{phan2022sleeptransformer}, and cross-modal flexibility \cite{Guo2024FlexSleepTransformer,Pradeepkumar2022CMT}. Foundation-style pretraining on full-night PSG also shows potential for robust downstream performance \cite{fox2025foundational} and interpretability via transformer explainability \cite{Horie2022SleepCAM,Lee2025SleepXViT}. Yet, most DL systems still approach sleep staging as a \emph{PSG-only} supervised problem, rarely leveraging contextual cues—such as expert event annotations (e.g., apneas, desaturations, arousals) or clinical metadata (age, sex, body mass index (BMI))—that human scorers use implicitly.

There are compelling physiological and empirical grounds to expect that incorporating expert annotations and clinical information would enhance sleep staging. Sleep architecture exhibits systematic variation with demographic and anthropometric factors—for instance, aging is associated with diminished N3 duration and altered REM expression \cite{li2022sleep}, while indices such as BMI affect sleep physiology and staging patterns through increased fragmentation and arousal susceptibility \cite{rao2009association,kim2022diet}. Respiratory disturbances are closely coupled to stage dynamics and cortical microstructure: approximately 90\% of cortical arousals linked to sleep-disordered breathing occur within about 20 s of the termination of a respiratory event \cite{zitting2023association}. Moreover, the AASM scoring guidelines explicitly connect respiratory effort–related arousals (RERAs) and hypopneas with cortical arousals or oxygen desaturations \cite{AASMManual2025,Malhotra2018Position}. In patients with OSA, both the arousal threshold and the burden of respiratory events shape sleep fragmentation and stage composition \cite{Younes2004,Sands2018}. Collectively, these findings indicate that per-epoch expert event labels and fundamental clinical attributes may provide complementary predictive signals beyond the PSG waveform itself.

Converging research lines support this perspective. Multi-task learning frameworks that simultaneously train on sleep staging and auxiliary objectives such as arousal or respiratory-event detection have demonstrated strong feasibility, suggesting a synergistic relationship between event cues and stage classification \cite{Zan2023FullSleepNet,Xie2024MTL}. Likewise, studies emphasizing robustness across varying OSA severities or incorporating disorder-specific information report improved staging performance when the model accounts for disease-related structure \cite{Lv2024SSleepNet}. Despite these encouraging results, the systematic fusion of \emph{expert per-epoch annotations} and \emph{clinical contextual signals} within modern transformer-based architectures for sleep staging remains largely uncharted.

This study aims to bridge that gap. We employ a vanilla transformer to derive per-epoch embeddings from PSG signals and subsequently integrate them across the night using a 1D convolutional head. Two complementary strategies are explored for contextual incorporation: (i) a \emph{feature-fusion} approach, in which subject-level clinical attributes (age, sex, BMI) and expert per-epoch annotation vectors (e.g., hypopnea, obstructive/central/mixed apnea, desaturation, respiratory-effort arousal, periodic breathing) are concatenated with transformer outputs prior to aggregation; and (ii) a \emph{multi-task} framework, where auxiliary prediction heads are jointly trained to classify these contextual variables alongside the primary sleep-stage objective.

Our experiments draw on the Sleep Heart Health Study (SHHS) cohort—merging SHHS-1 and SHHS-2 polysomnography data with standardized event annotations provided by the National Sleep Research Resource (NSRR)—to assess the impact of contextual integration in a large, heterogeneous sample \cite{Quan1997SHHS,zhang2024national,Kang2021SHHS125}. As detailed later, directly embedding expert and clinical information into the staging head yields consistent improvements over PSG-only baselines, whereas simply predicting the same variables through auxiliary tasks confers little additional benefit.

\section{Related Works}
\label{sec:related}

\subsection{From early CNNs to context-aware sequence models}
Early deep architectures for sleep staging on PSG treated each 30‑s epoch largely in isolation, often with lightweight temporal context. Representative CNNs operating on raw single‑channel EEG include the seminal works of Tsinalis \emph{et al.} and Sors \emph{et al.}, which demonstrated that learned time–frequency features surpass hand‑engineered ones across Sleep‑EDF and SHHS subsets \cite{tsinalis2016, sors2018}. Sequence‑to‑sequence hybrid designs then emerged to encode inter‑epoch dynamics—e.g., DeepSleepNet (CNN+BiLSTM) and SleepEEGNet (CNN+seq2seq), both showing sizeable gains in N1/REM recognition when context windows were used \cite{supratak2017, mousavi2019}. IITNet pushed this further by explicitly modeling intra‑epoch sub‑segments and inter‑epoch context using residual CNNs and BiLSTMs, improving macro‑F1 particularly for N1 \cite{seo2020}.

\subsection{Fully convolutional segmentation of full nights}
To alleviate the optimization difficulty and latency of recurrent models, Perslev \emph{et al.} reframed sleep staging as dense time‑series segmentation. U‑Time segments arbitrarily long sequences via a U‑Net‑like temporal FCN and aggregates at epoch resolution, yielding robust training and strong cross‑dataset performance \cite{Perslev2019UTime}. The successor U‑Sleep is a high‑frequency, fully convolutional system that scales across datasets and acquisition protocols and has become a common strong baseline in clinical benchmarking \cite{perslev2021usleep, fiorillo2023u}. Long‑range (multi‑cycle) temporal dependencies were later addressed by L‑SeqSleepNet, which efficiently encodes whole‑cycle ($\approx$ 90‑min) context and reports consistent gains across PSG and ear‑EEG variants \cite{phan2023lseqsleepnet}.

\subsection{Attention and Transformer families for PSG}
Attention mechanisms improved both accuracy and interpretability. AttnSleep combined convolutional feature extractors with attention to emphasize stage‑salient EEG patterns, especially for minority stages \cite{Eldele2021}. XSleepNet learned from multiple “views” (raw signals and spectrograms) with view‑adaptive training to improve complementarity and generalization \cite{phan2022xsleepnet}. Bringing self‑attention to the sequence level, SleepTransformer provided end‑to‑end sequence‑to‑sequence staging with interpretable attention maps and uncertainty quantification \cite{phan2022sleeptransformer}. 

\textbf{Multi‑channel PSG Transformers.} With growing interest in montage flexibility and richer cross‑channel context, several PSG‑only Transformer designs have been proposed. MultiChannelSleepNet uses Transformer encoders for per‑channel feature extraction and cross‑channel fusion and reports state‑of‑the‑art results on multi‑lead PSG \cite{dai2023multichannelsleepnet}. FlexSleepTransformer further introduces training on datasets with varying channel sets and demonstrates robustness to missing channels \cite{Guo2024FlexSleepTransformer}. Foundational/“full‑night” Transformers that encode entire nights at once have also been explored to reduce error propagation and improve global consistency \cite{fox2025foundational}.

\subsection{Montage‑agnostic training, transfer, and domain generalization}
Generalization across centers, montages, and demographics remains a key barrier to deployment. RobustSleepNet introduced a transfer learning strategy that is montage‑agnostic and demonstrated strong leave‑one‑dataset‑out performance across eight heterogeneous cohorts \cite{guillot2021robustsleepnet}. Earlier works explored personalization and transfer learning to adapt to subject‑specific characteristics or to transfer from large source cohorts to small targets, mitigating channel/device mismatch and class imbalance \cite{mikkelsen2018personalize, phan2020transfer}. Graph‑based or domain‑generalization approaches (e.g., multi‑view spatial–temporal GCNs) have likewise sought subject‑invariant features for PSG \cite{jia2021mstgcn}.

\subsection{Event‑ and context‑aware learning with expert information}
Clinical experts do not score stages in a vacuum: they consider micro‑events (K‑complexes, spindles, arousals) and respiratory events (apneas, hypopneas, desaturations) defined by AASM rules \cite{AASMManual2025,aasm_hypopnea_faq}. Deep event detection has matured (e.g., DOSED for spindles/K‑complexes/arousals, and top‑performing PhysioNet 2018 arousal detectors), providing high‑quality labels that can be coupled with staging \cite{chambon2019dosed, ghassemi2018you, warrick2018arousal, howe2018arousal}. Large epidemiologic PSG cohorts such as SHHS include expert‑annotated respiratory and arousal events \cite{Quan1997SHHS, redline1998shhs, zhang2024national}. Importantly, cortical arousals are tightly time‑locked to sleep‑disordered breathing events ($\approx$ 90\% occur within 20\,s of SDB termination), supporting the hypothesis that such event context is informative for stage boundaries and transitions \cite{zitting2023association}. While some prior works used structured output layers (HMM/CRF) to bias temporal consistency \cite{aggarwal2018neuralcrf}, a systematic injection of \emph{expert‑annotated event information} (e.g., apnea/hypopnea/arousal/desaturation/periodic breathing) directly into Transformer representations or via multi‑task heads has been less explored in the PSG‑only literature. This gap motivates our design: late fusion of epoch‑level event/clinical covariates with Transformer features and a comparison against multi‑task auxiliary prediction of the same information.

\subsection{Summary and positioning}
In summary, PSG‑only sleep staging progressed from epoch‑wise CNNs \cite{tsinalis2016, sors2018} to context‑aware hybrids \cite{supratak2017, mousavi2019, seo2020}, fully convolutional segmentation of full nights \cite{Perslev2019UTime, perslev2021usleep}, and Transformer families with improved interpretability and montage flexibility \cite{phan2022sleeptransformer, dai2023multichannelsleepnet, Guo2024FlexSleepTransformer}. Despite advances, two practical gaps persist: (i) leveraging \emph{expert‑annotated epoch metadata} (respiratory/arousal events) and basic \emph{clinical covariates} (age/sex/BMI) in a principled way alongside learned representations; and (ii) clarifying whether such information is better used as explicit inputs (conditioning) versus as auxiliary targets (multi‑task). We address both by (a) concatenating expert/clinical vectors to Transformer features before the temporal 1D convolution head, and (b) benchmarking a multi‑task alternative that jointly predicts stages and expert/clinical labels.

\section{Methods}

\subsection{Dataset Description}
We conducted our experiments using the SHHS dataset, a large multi-centre cohort of community-dwelling adults assessed for sleep disordered breathing and cardiovascular outcomes\cite{zhang2018national, Quan1997SHHS}. The SHHS collected in-home overnight PSG recordings from 6,441 participants at baseline (SHHS-1, 1995-1998) and a follow-up PSG from 3,295 of these individuals a few years later (SHHS-2, 2001-2003). We included both rounds. The recorded signals consisted of two bipolar EEG channels (C4-A1 and C3-A2), two EOG channels, one EMG channel, one ECG channel, thoracic and abdominal inductance plethysmography, a body position sensor, a light sensor, a pulse oximeter, and an airflow sensor. Sleep staging was performed manually by a single technician in 30-second epochs, classifying the data into Wake, N-REM (non-REM) stage 1, N-REM stage 2, N-REM stage 3, N-REM stage 4, and REM stage 5 stages. Participants with missing data in any of the required PSG channels (EEG, EOG, EMG, or cardiorespiratory signals) were excluded from the analysis to ensure signal completeness and comparability across subjects. After this exclusion, 8,357 subjects (86\% of the total) were retained.

\subsection{Data preprocessing}
The model input comprised nine signals: saturation of peripheral oxygen (SaO2), ECG, EMG, EOG (left and right), EEG, thoracic and abdominal respiratory signals, and body position. The EEG, ECG, EOG, and respiratory signals were subjected to band-pass filtering at $1.0-50.0$ Hz, $0.5-50.0$ Hz, $0.1-15.0$ Hz, and $0.0-1.0$ Hz, respectively. The original SHHS annotations were based on the classical Rechtschaffen and Kales (R\&K) scoring system \cite{Rechtschaffen1968Manual}, which differentiates N-REM stage 3 and stage 4 as separate deep-sleep stages. To harmonize label definitions with the current AASM standard, N-REM stages 3 and 4 (R\&K) were merged into a single N3 category prior to model training and evaluation. Other stages were mapped directly (Stage 1 → N1, Stage 2 → N2, REM → REM, Wake → Wake). Following the exclusion of recordings with incomplete channels, a final cohort of 8,357 subjects was established.

The dataset was partitioned into training, validation, and test sets with a subject-level split of 6,728, 788, and 841, respectively. For epoch-based analysis, each subject's polysomnography recording was segmented into non-overlapping 30-second epochs. This procedure yielded 7,264,099 epochs for the training set, 854,486 for the validation set, and 935,078 for the test set.

\subsection{Network structure}
The proposed model features a two-stage architecture, designed to sequentially perform intra-epoch feature extraction and inter-epoch contextual aggregation. The first stage consists of a Transformer-based encoder that processes individual epochs, while the second stage employs a 1D Convolutional Neural Network (CNN) to aggregate features across the entire sequence of epochs.

The Transformer encoder is composed of six stacked layers, composed with an eight-head multi-head attention mechanism and a feature dimension of 128. For the initial training of this encoder, an auxiliary Multi-Layer Perceptron \cite{Rumelhart1986} (MLP) head is appended. The feature vector for each epoch, generated by the Transformer, is flattened and fed into the MLP. This module is tasked with classifying the epoch into one of five sleep stages: Wake, N1, N2, N3, or REM.

The second stage, the 1D CNN aggregator, processes the sequence of epoch-level features extracted by the pre-trained Transformer. This aggregator comprises a stack of twelve 1D convolutional layers, each with a feature dimension of 512. The final convolutional layer reduces the feature dimension to 5, corresponding to the five sleep stages. This layer outputs the final class probability for each epoch by integrating contextual information from the entire sleep recording.

\subsection{Network training scheme}
Training scheme follows a sequential two-stage scheme designed to first learn robust epoch-level representations, and then learn to aggregate these representations along the epoch axis.

\paragraph{Stage 1: Intra-Epoch Feature Learning.}
The first stage focused on training the Transformer encoder. This stage utilized the complete set of 7,264,099 epochs from the training dataset. Each 30-second epoch served as an independent input to the Transformer, which, in conjunction with an appended MLP classification head, was trained to map the raw physiological signals to one of the five sleep stages (Wake, N1, N2, N3, REM). The objective of this stage was to learn the Transformer with the ability to extract salient, discriminative features from the time-series data within a single epoch.

\paragraph{Stage 2: Inter-Epoch Contextual Aggregation.} 
In the second stage, the pre-trained Transformer encoder from Stage 1 was utilized as a fixed feature extractor, with its weights frozen. The training shifted to the 1D convolutional network (CNN) aggregator, which learns to classify each epoch by considering the temporal context of the entire sleep session. For each subject, the sequence of all epochs was passed through the frozen Transformer to yield a sequence of feature vectors. To create uniform-sized inputs for the CNN, these feature sequences were standardized to a length of 1,500—the maximum sequence length in the cohort—by applying zero-padding to any shorter sequences. The zero-padded portions of the sequence were masked and excluded from this loss calculation, ensuring that only valid epochs contributed to the optimization. The aggregator was then trained end-to-end to predict the 1,500-length sequence of corresponding labels. The optimization was performed by applying a cross-entropy loss to the prediction for each epoch and summing the losses across the entire sequence.

\paragraph{Integration of Contextual Variables.} 
To investigate whether explicit clinical and event data could enhance staging performance, we integrated this information via a feature fusion mechanism. Subject-level clinical metadata (sex, age, etc.) and per-epoch annotation vectors (hypopnea, apnea, etc.) were concatenated with the output features from the Transformer for each respective epoch. This augmented feature sequence was then used as the input to the 1D CNN aggregator. To ablate the contributions of these information sources, we trained and evaluated four distinct models: (1) a baseline model using only PSG data; (2) a model augmented with subject-level metadata; (3) a model augmented with per-epoch annotations; and (4) a model utilizing both sources of contextual information.

\subsection{Training details}
The entire training process for all described schemes was implemented using PyTorch \cite{Paszke2019PyTorch}. The experiments were conducted with an NVIDIA L40S GPU and an Intel(R) Xeon(R) Gold 6448H CPU, with the full training procedure taking approximately 60 hours to complete. All network weights were initialized using the Xavier uniform initializer \cite{Glorot2010}. For optimization, we employed the Adam optimizer \cite{Kingma2014Adam} with a batch size of 16. The initial learning rate was set to $1 \times 10^{-4}$, and a learning rate decay with a factor of 0.90 was applied after each epoch. The models were trained by minimizing a cross-entropy loss function over a total of 100 epochs. To prevent overfitting and ensure generalization, the model checkpoint that achieved the lowest validation loss was selected as the final model for evaluation. The performance of the selected model on the test set was assessed using macro-F1 and micro-F1 scores.

\section{Results}
We evaluated our model's performance on the test set using Macro and Micro F1 scores, with the results summarized in Table 1. The baseline model, which relied solely on PSG signals, achieved a Macro F1 score of 0.7745. While the addition of subject-level clinical metadata yielded an improvement, a more substantial performance gain was observed when per-epoch event annotations were incorporated, boosting the Macro F1 score to 0.8018.

The optimal configuration, which integrated both clinical and event-related information, achieved the highest scores with a Macro F1 of 0.8031 and a Micro F1 of 0.9051. These findings demonstrate that while both types of contextual information are beneficial, the per-epoch event annotations provide the most significant contribution to the model's predictive accuracy.

\begin{table}[h!]
\centering
\caption{Sleep staging performance with different contextual information. The "None" configuration serves as the PSG-only baseline. "Clinical" refers to the addition of subject-level metadata, while "Event" refers to the addition of per-epoch expert event annotations.}
\label{tab:sleep_staging_performance}
\begin{tabular}{lcc}
\hline
\textbf{Model Configuration} & \textbf{Macro F1} & \textbf{Micro F1} \\ \hline
None (Baseline)        & 0.7745            & 0.8774            \\
+ Clinical             & 0.7811            & 0.8842            \\
+ Event              & 0.8018            & 0.9043            \\
+ Clinical \& Event    & \textbf{0.8031}   & \textbf{0.9051}   \\ \hline
\end{tabular}
\end{table}

Figure 1 visualizes the classification performance for all four model configurations using confusion matrices, complementing the aggregate F1 scores from Table 1 with a granular view of per-stage accuracy. The progression from the baseline (a) to the final model (d) visually confirms our findings: the addition of event annotations (c) significantly boosts accuracy, and the final model (d), integrating all contextual data, exhibits the strongest diagonal, aligning with its top-ranked F1 scores and demonstrating the highest overall accuracy.

\begin{figure}[h!]
\centering
\includegraphics[width=1\textwidth]{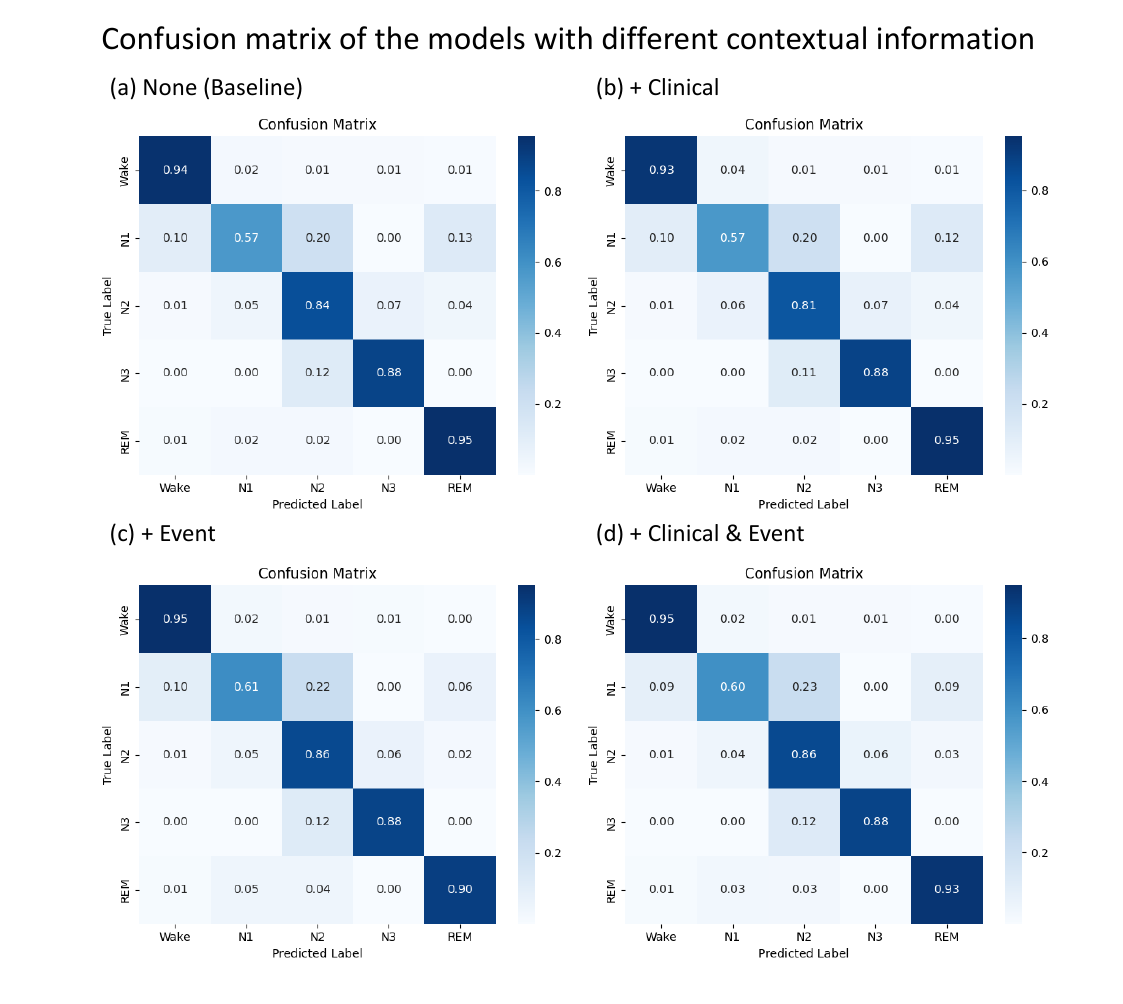}
\caption{Confusion matrices for sleep staging performance with different contextual information. (a) Baseline model (PSG-only), (b) model with clinical metadata (+ Clinical), (c) model with per-epoch event annotations (+ Event), and (d) the final model integrating all inputs (+ Clinical \& Event). The diagonal elements represent correctly classified epochs, visually demonstrating the improved accuracy as more information is added, particularly in model (d).}
\label{fig:confusion_matrices}
\end{figure}

In a separate ablation study, we investigated an alternative approach to leverage this contextual information. We explored a multi-task learning (MTL) framework by augmenting the baseline model with additional classification heads. These heads were tasked with predicting the subject-level clinical data and the per-epoch event annotations concurrently with the main sleep staging task. The performance of this approach compared to the baseline is detailed in Table 2.

\begin{table}[h!]
\centering
\caption{Ablation study comparing the baseline model with a multi-task learning (MTL) framework for integrating contextual information.}
\label{tab:mtl_ablation}
\begin{tabular}{lcc}
\hline
\textbf{Model Configuration} & \textbf{Macro F1} & \textbf{Micro F1} \\ 
\hline
None (Baseline) & 0.7745 & 0.8774 \\ 
MTL Approach & 0.7785 & 0.8790 \\ 
\hline
\end{tabular}
\end{table}

As shown in Table 2, the MTL approach (Macro F1: 0.7785) yielded only a negligible improvement over the baseline model (Macro F1: 0.7745). This joint training methodology did not provide the significant performance gains observed from direct feature integration (as shown in Table 1). These findings suggest that directly incorporating contextual information as input is the more effective integration strategy for this task.

\section{Discussion}
\subsection{Principal findings}
This study demonstrates that supplementing a PSG‑only transformer sleep staging pipeline with readily available clinical context (age, sex, BMI) and, critically, per‑epoch expert event annotations (e.g., apnea/hypopnea subtypes, desaturations, respiratory‑effort arousals, periodic breathing) yields consistent accuracy gains without altering the underlying montage or requiring external pretraining. On SHHS, macro‑F1 improved from 0.7745 → 0.8031 and micro‑F1 from 0.8774 → 0.9051 when both context sources were fused, with epoch‑level event annotations contributing the largest share of the improvement (macro‑F1 0.8018 with events alone). Feature fusion outperformed a multi‑task alternative that attempted to predict the same auxiliary information alongside staging. These results support a simple but clinically intuitive message: give the model the same cues human scorers use, and staging improves.

\subsection{Relation to prior work}
In recent years, transformers have emerged as the dominant backbone for PSG time-series analysis, enabling sequence-to-sequence sleep staging with uncertainty estimation and interpretability (SleepTransformer\cite{phan2022sleeptransformer}), as well as full-night foundation models that encode entire recordings to enhance global representations \cite{fox2025foundational}. Our findings complement, rather than contradict, these architecture-level advances: context injection provides additive improvements on top of a standard transformer and could be integrated with full-night encoders to further stabilize stage transitions.

Generalization to heterogeneous clinical settings is another active thread. Montage‑flexible transformers (e.g., FlexSleepTransformer\cite{Guo2024FlexSleepTransformer}) and transfer/domain‑generalization approaches (RobustSleepNet \cite{guillot2021robustsleepnet}; multi‑view spatiotemporal GCNs\cite{jia2021mstgcn}) address channel and cohort differences. Our late‑fusion design is orthogonal to these strategies: it leverages context that travels with the PSG (metadata and events) and should plug into montage‑agnostic or transfer‑learned backbones.

Finally, segmentation‑style systems (U‑Time/U‑Sleep\cite{Perslev2019UTime,perslev2021usleep}) and hierarchical sequence models (SeqSleepNet\cite{Phan2019SeqSleepNet}) established the value of long‑range temporal context; our two‑stage design (epoch encoder → whole‑night aggregator) adheres to this principle while showing that explicit clinical/event conditioning buys additional accuracy beyond what implicit temporal modeling alone provides.

\subsection{Physiological plausibility}
The gains from event conditioning are physiologically well‑motivated. Cortical arousals are tightly time‑locked to respiratory events: in a large clinical cohort, ~90\% of SDB‑associated arousals occur from ~6s before to ~14s after the event ends—exactly the windows where stage boundaries and microstructure are contentious. Feeding the model explicit information about apneas/hypopneas, desaturations, and effort‑related arousals therefore helps it resolve transition epochs and fragmented sleep\cite{zitting2023association}. 

Moreover, AASM definitions explicitly link hypopneas and RERAs to arousal/desaturation criteria, so event vectors encode rule‑level knowledge that complements learned waveform features\cite{AASMManual2025}. This mechanistic alignment likely explains why conditioning on provided events improved staging more than asking the network to infer them via auxiliary heads. 

Age, sex, and BMI also shape sleep architecture and fragmentation—e.g., less N3 and altered REM with aging and OSA‑related arousability modulated by the respiratory arousal threshold. Incorporating these factors at inference allowed the model to reflect subject‑specific priors that human scorers implicitly use\cite{li2022sleep, Sands2018}.

\subsection{Clinical and operational relevance}
Because the method does not change the montage, it is straightforward to deploy in labs that already capture the same channels and event annotations. In settings where automated tools are permitted, literature shows that automatic scoring can reduce technologist workload and variability while maintaining human‑level agreement—context‑aware models may amplify these benefits by being more attuned to clinically salient events\cite{Choo2023Benchmarking}.

Just as important, aligning model inputs with what clinicians consider (events + demographics) can improve interpretability: discrepancies between the hypnogram and event burden are easier to rationalize when both flow into the prediction. Emerging explainable transformer work suggests attention‑based visualizations and CAM‑like reasoning can make these dependencies visible to users\cite{phan2022sleeptransformer,Horie2022SleepCAM}.

\subsection{Strengths}
\begin{enumerate}
    \item \textbf{Large and heterogeneous cohort:} Utilizes the SHHS-1/2 datasets with standardized annotations from the NSRR, enhancing ecological validity and reproducibility.
    \item \textbf{Simple and reproducible fusion strategy:} Concatenating contextual information with epoch embeddings prior to the whole-night head yields measurable and consistent performance gains.
    \item \textbf{Clear ablation analyses:} Separately evaluates the effects of metadata and event annotations, highlighting the dominant influence of the latter.
\end{enumerate}

\subsection{Limitations}
\begin{enumerate}
    \item \textbf{Availability of event inputs at inference:} The largest gains arise from expert per-epoch event labels, which may be unavailable during real-time inference. Two practical remedies include: (i) pre-detecting events automatically and feeding those predictions as contextual inputs; or (ii) adopting multi-task architectures that jointly predict events and stages. Both strategies warrant prospective evaluation.
    \item \textbf{Guideline and site variability:} AASM scoring rules—particularly hypopnea definitions—evolve over time and vary across laboratories. External validation across AASM versions and scoring cultures remains essential. Encouragingly, recent studies indicate that high-capacity models (e.g., U-Sleep) exhibit resilience to such variability, though explicit verification is still required.
    \item \textbf{Montage and domain shift:} Although the proposed approach is montage-agnostic in principle, performance may degrade under changes in channel configuration or population distribution. Integrating montage-flexible architectures (e.g., FlexSleepTransformer) or employing transfer/domain-generalization techniques (e.g., RobustSleepNet, MSTGCN) alongside context fusion should be prioritized for cross-site deployment.
    \item \textbf{External validation:} While findings are derived from the large and diverse SHHS cohort, generalizability to external datasets with differing demographics, acquisition protocols, or annotation standards remains untested. Future work should confirm robustness across independent cohorts and clinical environments.
\end{enumerate}

\subsection{Future Directions}
\begin{enumerate}
    \item \textbf{Operational pipeline testing:} Evaluate a two-stage system (event detector $\rightarrow$ context-aware stager) within routine clinical workflows, quantifying both accuracy gains and time efficiency.
    \item \textbf{Integration with full-night encoders:} Explore whether combining foundation transformers with context-fusion mechanisms can mitigate boundary misclassifications (e.g., N1~$\leftrightarrow$~N2, N2~$\leftrightarrow$~REM).
    \item \textbf{Explainability and auditability:} Apply attention- or CAM-based visualization techniques to examine how detected events and metadata influence model decisions, ensuring that salient patterns correspond to AASM-defined hallmarks.
    \item \textbf{Version-robust training:} Incorporate AASM version or scoring-rule metadata as conditioning variables during training to address definition drift, and systematically report performance stratified by guideline version.
    \item \textbf{External validation:} Reproduce and extend the findings on independent datasets curated by the NSRR and other public or institutional cohorts to assess generalizability across diverse clinical settings.
\end{enumerate}

\section{Conclusion}
We proposed a Transformer-based sleep stage classification model that integrates expert-annotated knowledge, bridging deep learning with clinical insight. By jointly leveraging patient- and epoch-level annotations, our approach—unexplored in prior studies—significantly outperformed signal-only models in accuracy and F1-score. These expert-informed features proved crucial for challenging cases such as apnea-induced arousals and age-related sleep changes.

Our work advances automated sleep scoring toward human-expert performance while enhancing clinical interpretability. Future directions include automating expert feature generation, validating across external cohorts, and incorporating richer contextual data to further personalize predictions. In essence, this study demonstrates that coupling domain expertise with deep learning yields more accurate and clinically meaningful sleep staging—paving the way for next-generation intelligent sleep analysis systems.

\nocite{*}
\bibliographystyle{unsrt}
\bibliography{Reference/references}  
\end{document}